\documentclass[
aps,%
12pt,%
final,%
notitlepage,%
oneside,%
onecolumn,%
nobibnotes,%
nofootinbib,% In the current version of REVTeX, when this option is on,
superscriptaddress,%
noshowpacs,%
centertags]%
{revtex4}

\newtheorem{definic}{Definition}

\begin{document}
	%\selectlanguage{russian} % For a paper in Russian
	\selectlanguage{english} % For a paper in English
	\title[Increasing the throughput]
	{Increasing the throughput of machine translation systems using clouds}

	\author{\firstname{Jernej} \surname{Vi{\v c}i{\v c}}}
	%\thanks{}
	\affiliation{University of Primorska, Andrej Marusic Institute, Slovenia}
	\affiliation{Research Centre of the Slovenian Academy of Sciences and Arts, The Fran Ramov{\v s} Institute}
	\author{\firstname{Andrej} \surname{Brodnik}}
	%\thanks{}
	\affiliation{University of Primorska, Andrej Marusic Institute, Slovenia}
	\affiliation{University of Ljubljana, Faculty of Computer Science and Informatics, Slovenia}
	\begin{abstract}
The manuscript presents an experiment at implementation of a Machine Translation system in a MapReduce model. The empirical evaluation was done using fully implemented translation systems embedded into the MapReduce programming model. Two machine translation paradigms were studied: shallow transfer Rule Based Machine Translation and Statistical Machine Translation.

The results show that the MapReduce model can be successfully used to increase the throughput of a machine translation system. Furthermore this method enhances the throughput of a machine translation system without decreasing the quality of the translation output.

Thus, the present manuscript also represents a contribution to the seminal work in natural language processing, specifically Machine Translation. It first points toward the importance of the definition of the metric of throughput of translation system and, second, the applicability of the machine translation task to the MapReduce paradigm. 
	\end{abstract}
	\maketitle
\section{Introduction}
\label{introduction}
Most research in the area of machine translation evaluation focuses on the translation quality of the observed translation systems. 
The research presented in this manuscript focuses entirely on the throughput of translation system and proposes a method to increase the throughput with no effect on the quality of the translation.

There are quite a few cases where the machine translation throughput is a crucial aspect such as translation of large quantities of text, e. g. translating all the texts in the Project Gutenberg\footnote{Project Gutenberg Literary Archive Foundation: http://www.gutenberg.org/} or translating huge amounts of manuals in order to enter a new market, etc. Some of these cases can be solved using publicly available services such as Google translate\footnote{Google translate: http://translate.google.com/} or Microsoft Bing Translator\footnote{Microsoft Bing Translator: http://www.bing.com/translator} although the speed of translation (actually the amount of text that can be translated) is limited. However, there are cases where such approach is not viable, such as translating sensitive information ranging from local correspondence to proprietary literature or translating domain-specific texts where a proprietary translation system must be used. The obvious solution is using faster machines, but this solution requires new investment. Using public clouds like Amazon EC3 would reduce the investment costs, but for many applications the cost would still be too high. This approach would also involve an architecture change \cite{palankar2008}. Autodesk Brasil ventured in a one-time job of translating most of their manuals into Brazilian Portuguese, the job was done using the Apertium translation system as described in \cite{masselot2010}.

As said, the presented research focuses mainly on machine translation of large amounts of text on commodity machines (cost-effective). MapReduce is a programming model for processing big data sets in a distributed fashion. The basic question this research focuses on is how efficiently can a Machine Translation (MT) system be implemented in a MapReduce model? The translation task of large amounts of text can be divided into smaller units with no effect on the translation quality as all the machine translation systems base translations on independent translation units. Usually the translation of sentences is done independently although some research has been done on extending the boundaries for translation units over the sentence boundaries \cite{furuse1994}. The natural way of increasing the translation throughput (the number of translated words in a defined amount of time) is to translate parts of the text on separate translation systems as every sentence is translated independently.

The rest of the manuscript is organized as follows: The domain description is presented in sections \ref{mtse} through \ref{distributedcomputing}. The methodology is presented in section \ref{methodology}. The evaluation methodology with results is presented in section \ref{section:evaluation}. The manuscript concludes with the discussion and description of further work in section \ref{section:discussion}.

\section{Machine translation}
\label{mtse}
Machine translation as studied in this manuscript is an unsupervised process of translating from one natural language to another using computer programs.\footnote{European Association for Machine Translation: http://eamt.org/}

There has been almost no research on the topic of machine translation throughput mostly due to the fact that most of the research in the field of machine translation focuses on the machine translation quality and the throughput of the systems is at least a magnitude greater to the throughput of human translators. Some comparative research has been done examining the increase on the overall speed of translation process using machine translation tools compared to standard human translation process \cite{martinez2003} and also the effect of the using Computer Assisted Translation -- CAT tools that combine MT systems with translation memory and human post-editing process \cite{federico2012}.

\section{Machine translation throughput}
\label{mtt}
\begin{definic}
	\label{def:throughput}
	Translation throughput: $T=\frac{n}{t}$, where $n \equiv $ number of words in a text;$t \equiv $ setup time + translation time; \\
	Description: $T$ is the measured property of the translation system; $n$ is the number of units of the original text, in our case words; $t$ is the sum of the time needed to initialize the translation system and the amount of time the n words were translated.
\end{definic}

\begin{definic}
	\label{def:speedup}
	Increased speed of the translation throughput: $S=\frac{T_{new}}{T_{orig}}$; is the ratio between the new translation throughput and the original (reference) translation throughput.
\end{definic}

The evaluation requires a "big enough" testing sample that minimizes the startup effect to a desired minimum. We can rely on this simple metric because both translation paradigms base the translations on fixed-size chunks of text and both paradigms almost always ignore the global syntactic complexity of the sentences. This metric would not be fair if the experiment involved a system that parses the sentence.

The translation throughput as defined in Definition \ref{def:throughput} is the primary metric used in this manuscript for the performance evaluation (throughput) of the evaluated translation systems.

\section{Overview of machine translation systems}
\label{overview}
The following sections describe the translation system toolkits used in the experiment, both toolkits are often used opensource toolkits from the respective translation paradigms: 
\begin{itemize}
	\item Shallow-transfer Rule Based Machine Translation (Shallow Transfer RBMT) \cite{forcada1} paradigm that is most suited for translation of related languages \cite{vicic2009a}, represented by Apertium \cite{forcada2011};
	\item Statistical Machine Translation (SMT) \cite{alonazian1999,burbank2005} paradigm that is based on large quantities of data and mathematical models, represented by Moses \cite{koehn2007};
\end{itemize}

\subsection{Apertium}
\label{apertium}

\begin{figure}
	\centering
	\includegraphics[width=.70\textwidth]{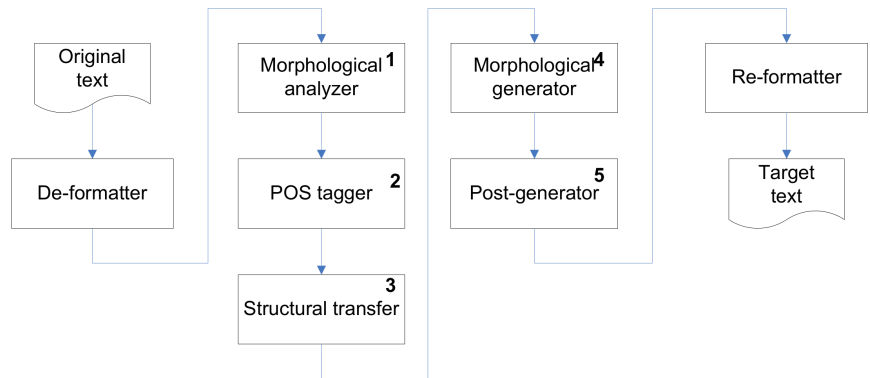}
	\caption{\small The modules of a typical shallow transfer RBMT translation system.}
	\label{fig:original_system}
\end{figure}

Apertium \cite{forcada2011} is an open-source machine translation platform, initially aimed at related-language pairs, but recently expanded to deal with more divergent language pairs (such as English -- Spanish). The shallow-transfer paradigm of the toolkit is best suited for related languages as the architecture does not provide the means for deep parsing which can lead to problems especially for the more divergent language pairs.
All these properties make Apertium a perfect choice for a cost effective development of a machine translation system for similar languages. The basic architecture of Apertium system is presented in Figure \ref{fig:original_system}. The systems \cite{forcada2011} and \cite{hajic2003} follow this design.

The numbered rectangles describe translation modules, output of a preceding module is the input to the successor:
\begin{enumerate}
	\item Morphological analyzer searches monolingual morphological dictionary of the source language to find all possible morphological tags and lemmata for the input word.
	\item POS tagger disambiguates the output of the preceding module by selecting the most probable tags.
	\item Structural and lexical transfer translates the disambiguated, morphologically analyzed text into the target language lexical units.
	\item Morphological generator searches the target dictionary for the appropriate word forms for the translated lexical units.
	\item Post-generator completes the automatic post-editing chores.
\end{enumerate}
Apertium is licensed under the LGPL.\footnote{GNU Lesser General Public License (LGPL)}

\subsection{Moses}
\label{moses}

Moses \cite{koehn2007} is recently the most widely used framework for setting up systems for statistical machine translation. The main features of the framework are:
\begin{itemize}
	\item Two types of conductive models (source models) based on phrases of the actual parts of the text (phrase-based), and based on trees (tree-based);
	\item To a certain extent permits the integration of explicit language knowledge at the level of words;
	\item Provides support for integration of tools with ambiguous outputs, such as morphosyntactic analyzers and parsers of speech and
	\item It is supported by large language models.
\end{itemize}

\begin{figure}
	\centering
	\includegraphics[width=.70\textwidth]{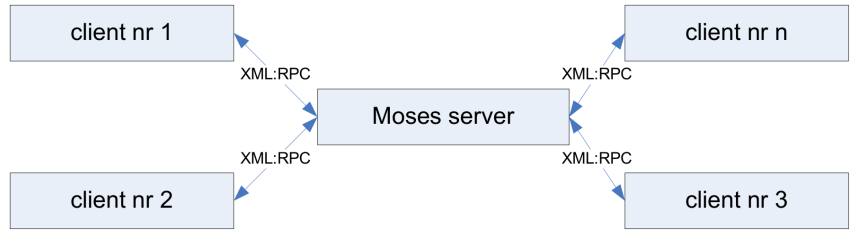}
	\caption{\small The Moses translation system can be deployed as a server, the communication is done through XML:RPC.}
	\label{fig:moses_system}
\end{figure}
Moses can be run in a server-like mode, where all the models are loaded into memory and the communication is done through XML:RPC. Figure \ref{fig:moses_system} shows a basic server deployment variation.
Moses is also licensed under the LGPL.\footnotemark[\value{footnote}]

\section{Related work}
\label{relatedwork}

There has been a considerable amount of research in speeding up the Machine Translation processes using parallel and distributed computing paradigms. Most of the work was done in the speeding up of the automatic learning processes in the area of the Statistical Machine Translation -- SMT \cite{dyer2009}. The most used Machine Translation toolkit, Moses \cite{koehn2007}, has already implemented support for multiple processor cores and to some extent for multiple computers.

A MapReduce-based large scale MT architecture has been proposed \cite{gao2008} which focuses on distributed storage for streaming and structured data that could be employed, the proposed architecture mainly focuses on the SMT paradigm. Training phase for SMT based on MapReduce has been proposed by \cite{dyer2009}.

The translation phase received less attention from the research community although there were a few successful attempts such as \cite{chao2010} using GPUs and focusing on the SMT paradigm.

The MapReduce paradigm was used as a speedup tool for the automatic translation of texts by \cite{rashidahmad2011}; their work focuses on one Rule-Based Machine Translation system.

The Apache Hadoop \cite{white2012} framework transparently provides both reliability and data delivery to applications -- moving data to processors (computers) that do task execution in contrast to systems such as BOINC \cite{marosi2013} or HTCondor \cite{thain2005}. 

The main contribution of this manuscript in comparison to the related work is the inclusion of two most popular MT paradigms (RBMT and SMT) and the focus on accelerating the translation phase. The main deficiency of the aforementioned systems (BOINC \cite{marosi2013} or HTCondor \cite{thain2005}) in comparison to Hadoop is their lack of support for data movement. This is provided in Hadoop through the Hadoop Distributed FileSystem. From this perspective one shall understand SMT as a service in a system that is installed on individual nodes in a similar way as any other service.

\section{Distributed computing}
\label{distributedcomputing}

In general, distributed systems are used to solve hard and parallel computational problems. In distributed computing, a problem is divided into many tasks, each of which is solved by one or more computers \cite{tanenbaum2006} that communicate with each other by message passing \cite{andrews2000}. To increase the efficiency it is desired to minimise the need to exchange information between the tasks.
Main advantages of distributed computing are: 
\begin{itemize}
	\item Users are distributed;
	\item Information is distributed;
	\item It may be more reliable if used correctly and
	\item It may be faster and cheaper, especially in comparison with supercomputers.
\end{itemize}
The main conceptual problem is the disassociating of a task and its data: data are on one computer while the task is on another.
The main technical problems are: 
\begin{itemize}
	\item Computers are prone to malfunctions, increasing the number of computers also increases the probability of failure;
	\item Network connections are falling;
	\item Data transmission is slow (10Gbit network has 300 micro-second latency \cite{lua2005}) 2GHz processor does 600,000 cycles in the same amount of time and
	\item There is no global clock (ticks) for the whole system.
\end{itemize}
These problems are also addressed by the MapReduce \cite{dean2010} programming model.
\subsection{MapReduce and Hadoop}
\label{mapreduce}
MapReduce \cite{dean2010} is a programming model for processing large data sets, and the name of an implementation of the model by Google. The model was developed by Google from the Map-fold model which has its roots in functional programming. It eases communication and coordination, rescue of crashed computers (moving load to available nodes), status reporting, debugging and basic optimization.

\begin{figure}
	\centering
	\includegraphics[width=.70\textwidth]{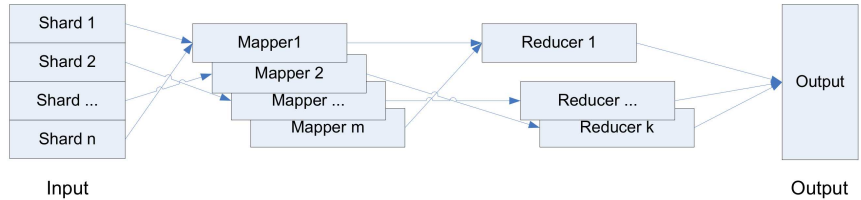}
	\caption{\small MapReduce setting, two basic phases: mapping and reducing in parallel.}
	\label{fig:mapreduce}
\end{figure}

The basic architecture of a MapReduce setting is shown in Figure \ref{fig:mapreduce}, shards are basic data chunks, mappers extract information from data shards and feed the extracted information to the Reducers that accumulate or generalize the results.

Apache Hadoop \cite{white2012} is an open-source software framework that supports data-intensive distributed applications. It supports running applications on large clusters of commodity hardware. Hadoop is based on Google MapReduce \cite{dean2010} and Google File System (GFS) \cite{ghemawat2003}.
The Hadoop framework transparently provides both reliability and data delivery to applications -- moving data to processors (computers) that do task execution in contrast to BOINC \cite{marosi2013} or HTCondor \cite{thain2005} systems ref. Figure \ref{fig:mapreduce}. 
Hadoop is licensed under the Apache license.\footnote{Apache v2 license}

\section{Methodology}
\label{methodology}
The research presented in this manuscript focuses entirely on the throughput of the translation system and proposes a method to increase the throughput with no effect on the translation quality. The throughput of the translation is measured as in Definition \ref{def:throughput}.

To employ Machine Translation in a MapReduce model, we must first find parallelism in our data and/or algorithms. The data parallelism comes from the fact that sentences are independently translated. The first assumption for the shard length can be a sentence. Finding parallelism in translation algorithms is beyond the scope of this research.

The movement of data to the processing nodes is in Hadoop provided through the Hadoop Distributed FileSystem. From this perspective one shall understand SMT as a service in a system that is installed on individual nodes in a similar way as any other service. Now, the question is how to translate massive amounts of data. The data is split into shards and each of them is mapped to one client machine -- the map phase. The translations are then collected into a final translation -- the reduce phase. 

The two most commonly used machine translation paradigms were addressed in the experiment, each paradigm was presented by an open-source solution. The SMT \cite{alonazian1999,burbank2005} solution was  Moses \cite{koehn2007}, the RBMT solution was Apertium \cite{forcada2011}.

The language pair for the Apertium system was English -- Spanish (EN -- ES), which is available under GPL license.\footnote{https://svn.code.sf.net/p/apertium/svn/trunk/apertium-en-es} The same language pair was used for Moses to enable the reuse of the same test-data.

\section{Evaluation methodology and results}
\label{section:evaluation}

The setting for the experiment involved constructing test data, deploying the translation system and measure the time the system needed to translate the prepared test set. Three basic objectives were sought:
\begin{itemize}
	\item Elimination of the startup time effect;
	\item Evaluation of the translation throughput of the translation system on one machine;
	\item Evaluation of the translation throughput of the translation system using a MapReduce cluster.
\end{itemize}

\subsection{Test setting}
\label{section:testsetting}

The test environments were installed on a cluster of commodity machines\footnote{\label{slow}Pentium(R) Dual-Core CPU E5300@2.60GHz, 4GB RAM, Gigabit ethernet.} that were used as the main testing deployment and on a faster machine\footnote{\label{fast}Intel(R) Core(TM) i7-3930K@3.20GHz, 32GB RAM, Gigabit ethernet.} that was used as a reference. The Apertium uses algorithms that are almost independent of the input text when regarding only translation throughput. The same fact can be attributed to the algorithms used by the Moses system if the caching option is turned off.

The operating system on all test machines was Ubuntu 14.04 LTS (Trusty Tahr), the only difference was that Server version was installed on the fast machine and Desktop version on the cluster machines.

The version of Apertium used in the experiments was 3.2.
The version of Moses used in the experiments was 2.1.1. The system was trained on the Europarl v7 corpus \cite{koehn2005}. The system was used with all default switches except the caching option was turned off. All the translation data was binarised.

\subsection{Test data}
\label{section:testdata}
The main test data set was artificially constructed in order to eliminate the possible effects of the non-uniformity of the test data. The artificial sentences were constructed from one sentence composed of 20 words which were present in the dictionary and copied the desired number of times. The sentence length was chosen as an approximate upper bound mean value of the sentence length in Opus corpus \cite{tiedemann2012} (the exact value is $16.5$) and as an upper limit of the sentence length in Google n-gram corpus (the exact value is $10.8$) \cite{google2006}. This data set would be a problematic selection if the quality of the translation was measured, or if complex parsing algorithms were involved in the translation. The selection of simple translation techniques (Apertium) and mathematical models (Moses) allows the usage of artificial test data. The only problem arising using this simple test-data set was the caching option adopted by Moses which caches previous translations and could benefit in throughput greatly from this feature. The system was tested using the same test set and same test setting with this option turned on and off. The results presented in Table \ref{tab:mosescache} show a significant influence of the caching to the artificial test data although the time differences are linear to the amount of input data. All further tests were done with the caching feature turned off.

\begin{table}[h]
	\centering
	\caption{\small The influence of caching on translation throughput in Moses. The caching significantly influences the throughput of artificial data.}
	\begin{tabular}{rrcrr}
		\cline{1-5}
		\cline{1-5}
		Nr. words & Nr. sent. & System & Real time & words/s\\
		\cline{1-5}
		2,000 & 100 & seq. moses(fast)-caching ON& 0:51.11 & 39.1\\
		20,000 & 1,000 & seq. moses(fast)-caching ON& 8:22.26 & 39.8\\
		2,000 & 100 & seq. moses(fast)-caching off & 1:22.89 & 24.1\\
		20,000 & 1,000 & seq. moses(fast)-caching off & 13:44.39 & 24.3\\
		\cline{1-5}
	\end{tabular}
	\label{tab:mosescache}
\end{table}

The possible influences of the artificial test-set were further observed by including a real-life data test set \cite{specia2011}. Most of the results are presented for both test sets.

All the test data is publicly available to facilitate the re-execution of experiments at the Language technologies server of the University of Primorska.\footnote{Test data: \url{http://jt.upr.si/research_projects/mapreduce_mt_throughput/}}

\subsection{Sequential system}
\label{section:sequentialsystem}

The first experiment involved measuring the throughput of the standard installations. Two sequential settings were deployed, one for each translation toolbox (Apertium and Moses). The translation systems were installed on the same set of machines (one testing\textsuperscript{\ref{slow}} and a reference machine\textsuperscript{\ref{fast}}) and the translation throughput was tested using the same test-sets. Both settings were tested using different sizes of source texts. 

The results of the Apertium system using differently sized artificial test data are presented in table \ref{tab:testdataapertium}. It shows the test data set with the results of the evaluation of the translation throughput of a single system. The throughput is in words per second (using real time). A steep increase of throughput using a small number of sentences which can be attributed to a fixed setup time and a linear time spent for each sentence. The exact setup time cannot be measured as the translation pipeline starts in parallel, succeeding pipeline stages are waiting for the output of the preceding stages. The influence of the startup time is not significant when translating more than 10,000 sentences or 200,000 words in our case. The translation throughput stabilizes at around 4,500 words per second on the test environment machine. The results on the reference machine\textsuperscript{\ref{fast}} are almost perfectly linear (twice as fast) for all tests.

\begin{table}[h]
	\centering
	\caption{\small Translation time on a single-computer setting (both hardware variants) for Apertium architecture using artificial test data (translating from Spanish).}
	\begin{tabular}{rrcrr}
		\cline{1-5}
		Nr. of words & Nr. of sent. & System & Real time & words/s\\
		\cline{1-5}
		2,000 & 100 & seq. apertium & 00:01.78  & 1,124\\
		20,000 & 1,000 & seq. apertium & 00:05.24  & 3,817\\
		200,000 & 10,000 & seq. apertium & 00:44.85  & 4,459\\
		2,000,000 & 100,000 & seq. apertium & 08:37.82 & 4,672\\
		\cline{1-5}
		2,000 & 100 & seq. apertium-fast & 00:00.91  & 2,198\\
		20,000 & 1,000 & seq. apertium-fast & 00:02.60  & 7,692\\
		200,000 & 10,000 & seq. apertium-fast & 00:20.22  & 9,891\\
		2,000,000 & 100,000 & seq. apertium-fast & 03:16.36 & 10,185\\
		\cline{1-5}
	\end{tabular}
	\label{tab:testdataapertium}
\end{table}

The results of the Apertium system using the real-life data test set \cite{specia2011} (the whole set and the same text copied twice) are presented in Table \ref{tab:testdataapertiumrealdata}. This test was used to show the possible influences of the artificial test-set on the results.

\begin{table}[h]
	\centering
	\caption{\small Translation time on a single-computer setting for Apertium architecture using real life test data (translating from Spanish).}
	\begin{tabular}{rrcrr}
		\cline{1-5}
		\cline{1-5}
		Nr. of words & Nr. of sent. & System & Real time & words/s\\
		\cline{1-5}
		21,118 & 1,000 &seq. apertium& 00:06.92  & 3,052\\
		42,236 & 2,000 &seq. apertium& 00:10.04  & 4,207\\
		\cline{1-5}
		21,118 & 1,000 &seq. apertium-fast& 00:03.03  & 6,970\\
		42,236 & 2,000 &seq. apertium-fast& 00:05.30  & 7,969\\
		\cline{1-5}
	\end{tabular}
	\label{tab:testdataapertiumrealdata}
\end{table}

Table \ref{tab:testdatamoses} shows the test data with the results of the evaluation of the translation throughput of a single system. The not applicable label (na\footnote{\label{na}Some of the long-running tests (many days) were skipped due to time constraints.}) denotes the long-running tests (many days) that were skipped due to time constraints. The system \emph{sequential - moses} was deployed on the same computer as the system shown in Table \ref{tab:testdataapertium}. 

The system \emph{sequential - moses (fast)} was deployed on a faster computer\textsuperscript{\ref{fast}}. This system is used as a reference system to show how fast we can get with much better hardware (meaning higher costs), the system was not used in MapReduce experiments. The throughput is in words per second (using real time). All the other parameters of the experiment are the same as in the experiment presented in Table \ref{tab:testdataapertium}. The throughput stabilizes at roughly $11$ words/s on the standard computer and roughly twice as much ($23.6$) on the faster reference computer.

\begin{table}[h]
	\centering
	\caption{\small Test data and translation time on a single-computer setting for Moses architecture.}
	\begin{tabular}{rrcrr}
		\cline{1-5}
		\cline{1-5}
		Nr. of words & Nr. of sent. & System & Real time & words/s\\
		\cline{1-5}
		200 & 10 & seq. moses & 0:0:16.83 & 11.9\\
		2,000 & 100 & seq. moses& 0:2:46.04 & 12.0\\
		20,000 & 1,000 & seq. moses & 0:33:22.39 & 10.0\\
		200,000 & 10,000 & seq. moses & 5:14:59.74 & 10.58\\
		2,000,000 & 100,000 & seq. moses & na\textsuperscript{\ref{na}} & na\textsuperscript{\ref{na}}\\
		\cline{1-5}
		200 & 10 & seq. moses(fast) & 0:0:08.50 & 23.5\\
		2,000 & 100 & seq. moses(fast) & 1:22.89 & 24.1\\
		20,000 & 1,000 & seq. moses(fast) & 13:44.39 & 24.3\\
		200,000 & 10,000 & seq. moses(fast) & 2:21:24.10 & 23.6\\
		2,000,000 & 100,000 & seq. moses(fast) & na\textsuperscript{\ref{na}} & na\textsuperscript{\ref{na}}\\
		\cline{1-5}
	\end{tabular}
	\label{tab:testdatamoses}
\end{table}

Table \ref{tab:testdatamosesrealdata} presents the same system using the \cite{specia2011} data set. This test was used to show the possible influences of the artificial test-set on the results.

\begin{table}[h]
	\centering
	\caption{\small Translation time on a single-computer setting for Moses architecture using real life test data (translating from Spanish).}
	\begin{tabular}{rrcrr}
		\cline{1-5}
		\cline{1-5}
		Nr. of words & Nr. of sent. & System & Real time & words/s\\
		\cline{1-5}
		21,118 & 1,000 &seq. moses& 25:04.95  & 14.0\\
		42,236 & 2,000 &seq. moses& 50:12.32  & 14.0\\
		\cline{1-5}
		21,118 & 1,000 &seq. moses-fast& 12:18.93  & 28.6\\
		42,236 & 2,000 &seq. moses-fast& 24:55.78  & 28.2\\
		\cline{1-5}
	\end{tabular}
	\label{tab:testdatamosesrealdata}
\end{table}

The throughput values are consistent using both test sets which shows that the artificial test set does not influence the results (if the caching option is turned off in Moses system).

\subsection{Distributed system}
\label{section:distributedsystem}

Three architectures were implemented in the MapReduce model. The actual translation was done in the mapping phase of the MapReduce model in all three architectures.

\begin{figure}[h]
	\centering
	\includegraphics[width=.75\textwidth]{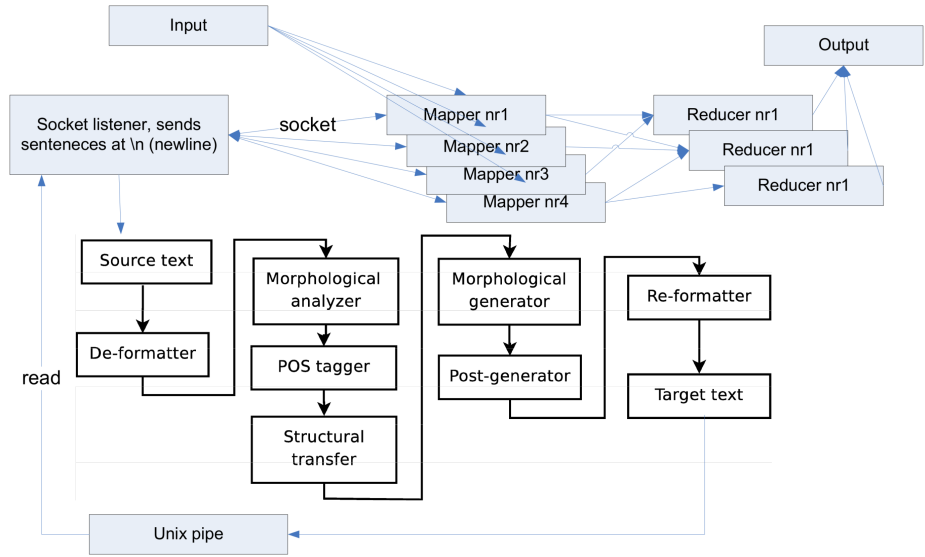}
	\caption{\small The architecture that minimizes the startup time impact.}
	\label{fig:newpipeline}
\end{figure}

The first architecture (Apertium service architecture) employed a simple service in order to minimize the startup effect of the translation system. Figure \ref{fig:newpipeline} shows the architecture with a service that communicates with mappers through sockets and with the translation system through POSIX pipes. A semaphore provides a locking mechanism to ensure data integrity. The server services mappers in a FIFO style. Communication is one-way; mappers simply deliver the data to the server and continue.

The presented architecture minimized the startup effect, but the communication overhead was quite substantial. The MR-service-apertium system in Table \ref{tab:serviceresults} presents empirical evaluation of the service architecture in the MapReduce environment. 

\begin{table}[h]
	\centering
	\caption{\small The comparison of the MapReduce implementation of the Apertium simple and service architecture. The service system is faster for smaller and slower for larger shards of input data. The break even point is 500 sentences (10,000 words).}
	\begin{tabular}{rrcrrr}
		\cline{1-6}
		\cline{1-6}
		Words & Sent. & System & Real time & Nodes& Words/s\\
		\cline{1-6}
		200 & 10 & MR-service-apertium & 03:21,82 & 1 & 1.0\\
		2,000 & 100 & MR-service-apertium & 03:22,40 & 1 & 9.9\\
		20,000 & 1,000 & MR-service-apertium & 03:27,75 & 1 & 96.3\\
		200,000 & 10,000 & MR-service-apertium & 04:25,36 & 1 & 753.7\\
		2,000,000 & 100,000 & MR-service-apertium & 14:04,71 & 1 & 2367.7\\
		\cline{1-6}
		200 & 10 & MR-simple-apertium & 03:21,37 & 1 & 1.0\\
		2,000 & 100 & MR-simple-apertium & 03:22,45 & 1 & 9.9\\
		20,000 & 1,000 & MR-simple-apertium & 03:27,16 & 1 & 96.5\\
		200,000 & 10,000 & MR-simple-apertium & 04:10,32 & 1 & 799.0\\
		2,000,000 & 100,000 & MR-simple-apertium & 11:24,88 & 1 & 2920.2\\
		\cline{1-6}
	\end{tabular}
	\label{tab:serviceresults}
\end{table}

The break even point was 500 sentences, meaning that the new architecture was faster translating less that 500 sentences in one job and it was slower than the original architecture for bigger jobs. We decided to eliminate the startup effect by setting the shard size at 1,000 sentences (well above the break-even point) and so minimizing the startup effect.

The service architecture was discarded for the simpler architecture with larger shards of input data (20,000 words). Table \ref{tab:finalresults} shows the results of the evaluation, the MR-apertium system is the same as MR-simple-apertium for each node of the MR setting. The throughput of the translation system is almost linear to the number of nodes in the cluster. The break even point for the new setting is 4 nodes, meaning the MapReduce installation is faster with only four nodes and the throughput increases almost linearly to 16 which was the maximum number of nodes used in our experiment.

\begin{figure}
	\centering
	\includegraphics[width=.70\textwidth]{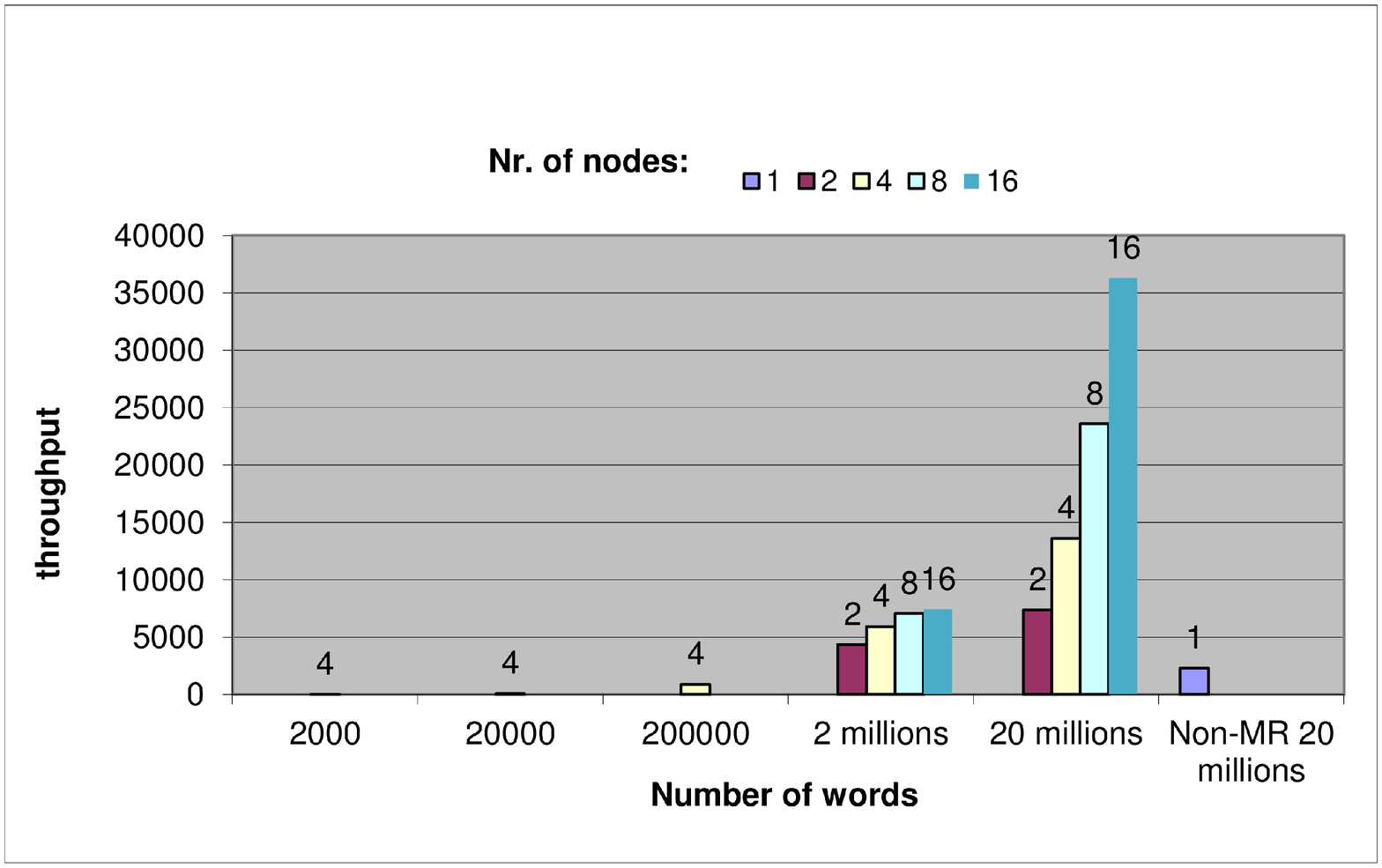}
	\caption{\small The final results for RBMT system (Apertium): the throughput is linear to the number of the nodes in the cluster ($n_{max} = 16$) providing there is enough data to translate (20,000,000 words).}
	\label{fig:final_apertium}
\end{figure}

\begin{table}[h]
	\centering
	\caption{\small The final results for RBMT system (Apertium): the throughput is linear to the number of the nodes in the cluster ($n_{max} = 16$) providing there is enough data to translate (20,000,000 words).}
	\begin{tabular}{rrcrrr}
		\cline{1-6}
		\cline{1-6}
		Words & Sent. & System & Real time & Nodes& Words/s\\
		\cline{1-6}
		2,000	&	100	&	MR-apertium	&	00:03:32.63	&	4	&	9.4\\
		20,000	&	1,000	&	MR-apertium	&	00:03:34.75	&	4	&	93.1\\
		200,000	&	10,000	&	MR-apertium	&	00:03:45.15	&	4	&	886.3\\
		2,000,000	&	100,000	&	MR-apertium	&	00:07:28.09	&	2	&	4,362.9\\
		2,000,000	&	100,000	&	MR-apertium	&	00:05:32.75	&	4	&	5,918.6\\
		2,000,000	&	100,000	&	MR-apertium	&	00:04:40.99	&	8	&	7,052.6\\
		2,000,000	&	100,000	&	MR-apertium	&	00:04:28.00	&	16	&	7,426.8\\
		20,000,000	&	1,000,000	&	MR-apertium	&	00:43:27.82	&	2	&	7,377.3\\
		20,000,000	&	1,000,000	&	MR-apertium	&	00:23:39.04	&	4	&	13,597.2\\
		20,000,000	&	1,000,000	&	MR-apertium	&	00:13:42.02	&	8	&	23,588.5\\
		20,000,000	&	1,000,000	&	MR-apertium	&	00:08:58.23	&	16	&	36,285.8\\
		\cline{1-6}
	\end{tabular}
	\label{tab:finalresults}
\end{table}

\begin{figure}[h]
	\centering
	\includegraphics[width=.75\textwidth]{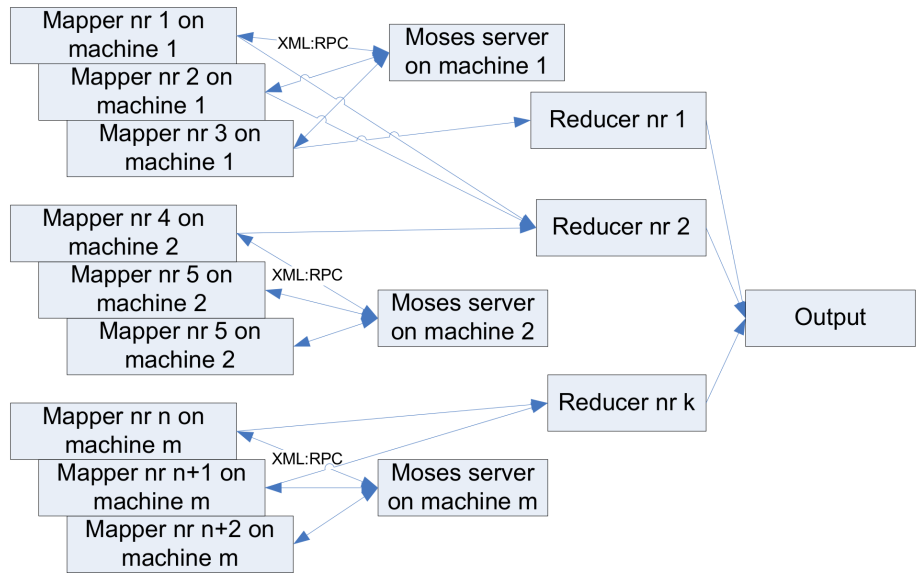}
	\caption{\small The architecture for the Moses MapReduce implementation.}
	\label{fig:mosespipeline}
\end{figure}

The Moses framework already has a server deployment option that was used in the MapReduce implementation of the Moses-based translation system in the experiment. Servers were installed on each node to minimize the network traffic. Figure \ref{fig:mosespipeline} shows the architecture used for the experiment. Translation is done in the Map phase using an XML:RPC call to the Moses server residing on the same physical node. 

The comparison of the impact on the throughput using the Moses server deployment and XML:RPC Java client is presented in Figure \ref{tab:mosesserver}. The results show only a marginal throughput loss. The setting was tested on the reference (fast) machine.

\begin{table}[h]
	\centering
	\caption{\small The impact on the throughput using Moses server and XML:RPC Java client.}
	\begin{tabular}{rrcrr}
		\cline{1-5}
		\cline{1-5}
		Words & Sent. & System & Real time & Words/s\\
		\cline{1-5}
		200	&	10	&	mosesserver (fast)	&	0:0:8.7 & 23.0\\
		2,000	& 100	& mosesserver	(fast)	&	0:1:23.98 & 23.8\\
		20,000	& 1000	& mosesserver	(fast)	&0:14:12.35 & 23.5\\
		200,000	& 10,000	& mosesserver	(fast)	&	2:21:45.87 & 23.5\\
		2,000,000	& 100,000	& mosesserver	(fast)	&	na\textsuperscript{\ref{na}} & na\textsuperscript{\ref{na}} \\
		\cline{1-5}
	\end{tabular}
	\label{tab:mosesserver}
\end{table}

\begin{figure}
	\centering
	\includegraphics[width=.70\textwidth]{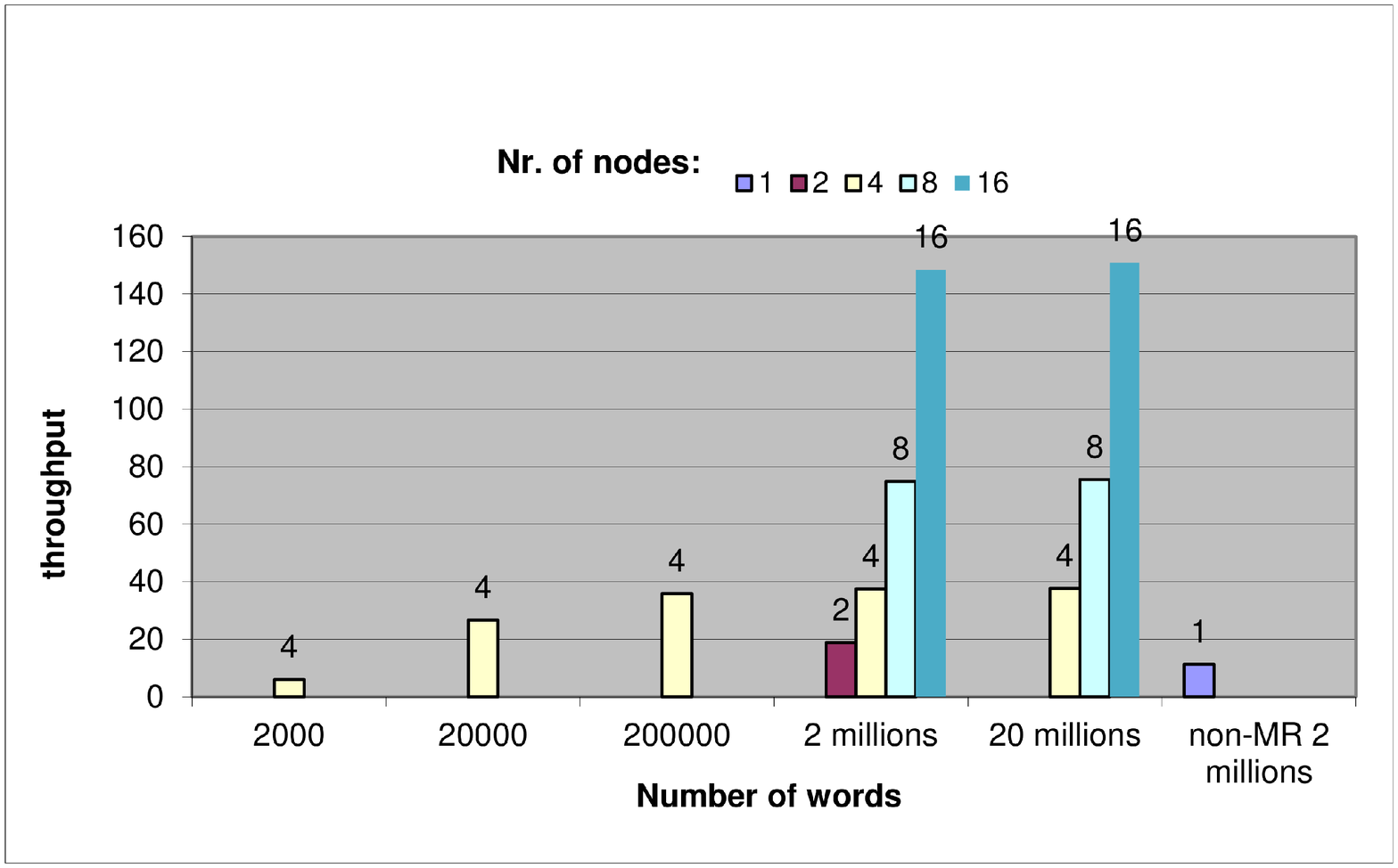}
	\caption{\small The final results for SMT system: the throughput is linear to the number of the nodes in the cluster (n\_max = 16).}
	\label{fig:final_moses}
\end{figure}

The results of the evaluation are presented in Table \ref{tab:finalresultsmoses} and in Figure \ref{fig:final_moses}. The throughput is linear to the number of nodes, the penalty for using the Map reduce setting is the startup time of around 3 minutes and a general $20 \%$ lower throughput due to the more complicated architecture.

\begin{table}[h]
	\centering
	\caption{\small The final results for SMT system: the throughput is linear to the number of the nodes in the cluster (n\_max = 16).}
	\begin{tabular}{rrcrrr}
		\cline{1-6}
		Words & Sent. & System & Real time & Nodes& Words/s\\
		\cline{1-6}
		2,000	 & 	100	 & 	MR-moses	 & 	0:4:20.37	 & 	4	 & 	6.0\\
		20,000	 & 	1,000	 & 	MR-moses	 & 	0:13:32.24	 & 	4	 & 	26.7\\
		200,000	 & 	10,000	 & 	MR-moses	 & 	1:38:27.44	 & 	4	 & 	35.9\\
		2,000,000	 & 	100,000	 & 	MR-moses	 & 	31:13:11.16	 & 	2	 & 	18.8\\
		2,000,000	 & 	100,000	 & 	MR-moses	 & 	15:40:28.55	 & 	4	 & 	37.6\\
		2,000,000	 & 	100,000	 & 	MR-moses	 & 	7:51:24.39	 & 	8	 & 	74.9\\
		2,000,000	 & 	100,000	 & 	MR-moses	 & 	3:58:1.17	 & 	16	 & 	148.4\\
		20,000,000	 & 	1,000,000	 & 	MR-moses	 & 	na	 & 	2	 & 	na\\
		20,000,000	 & 	1,000,000	 & 	MR-moses	 & 	156:42:18.55	 & 	4	 & 	37.8\\
		20,000,000	 & 	1,000,000	 & 	MR-moses	 & 	78:20:3.69	 & 	8	 & 	75.5\\
		20,000,000	 & 	1,000,000	 & 	MR-moses	 & 	39:1:47.45	 & 	16	 & 	150.8\\
		\cline{1-6}
	\end{tabular}
	\label{tab:finalresultsmoses}
\end{table}

\section{Discussion and further work}
\label{section:discussion}
The aim of the experiment was to test if the MapReduce model is suitable for machine translation tasks. The most used open source toolbox was chosen for each of the two most popular translation system paradigms, RBMT and SMT. The systems were tested in a MapReduce model. It was empirically proven, that the MapReduce programming model is suitable for machine translation task after architectural combination of Hadoop and individual MT systems.

The increase in throughput for the presented RBMT system was roughly 800 \% using the 16 machines in the testing cluster over the single machine (one of the machines in the cluster). The shards for the translation task were 1,000 sentences or more.

The increase in throughput for the presented SMT system was roughly 1,200 \% using the 16 machines in the testing cluster over the single machine (one of the machines in the cluster). The experiments will be repeated on a larger cluster as the empirical results show almost linear increase in the translation throughput by increasing the number of nodes in the cluster (our limit in the test was 16).

Further work can be done searching for parallelism not only in data, but also in algorithms although this would mean basing the research on a single MT paradigm and using MapReduce paradigm there. Additional research should be done searching for a further limitation of the setup time effect on the overall performance since we selected a simplistic approach using bigger shards.

\bibliographystyle{maik}
\bibliography{library}

\end{document}